# Decision-Theoretic Control of Problem Solving: Principles and Architecture

**John S. Breese**
**Michael R. Fehling**
Rockwell International Science Center
Palo Alto Laboratory
444 High Street
Palo Alto, CA 94301

July 1988

## Abstract

This paper presents an approach to the design of autonomous, real-time systems operating in uncertain environments. We address issues of problem solving and reflective control of reasoning under uncertainty in terms of two fundamental elements: 1) a set of decision-theoretic models for selecting among alternative problem-solving methods and 2) a general computational architecture for resource-bounded problem solving. The decision theoretic models provide a set of principles for prioritizing the assignment of computational resources among multiple problem-solving activities under uncertainty and with respect to various time constraints. Alternative problem-solving methods are chosen based on their relative costs and benefits, where benefits are characterized in terms of the value of information provided by the output of a reasoning activity. The output may be an estimate of some uncertain quantity or a recommendation for action. The computational architecture, called Schemer-II, supports the interleaving of, and communication among, various problem-solving subsystems that provide alternative approaches to information gathering, belief refinement, solution construction, and solution execution. We discuss the role of decision-theoretic control in an architecture such as Schemer-II for scheduling problem-solving elements and for critical-event-driven interruption of activities.

## 1 Introduction

An autonomous system, operating in a complex and constantly changing environment, must formulate and carry out plans to achieve desired behaviors or objectives. In such an environment the synthesis and use of plans will typically be severely constrained by limitations on time, information, and other critical resources. In response to these dynamically changing constraints, the system must be able to judiciously manage its reasoning and other activities to make the best use of available resources. We refer to problem solving under these conditions as resource-bounded problem solving: controlling and adapting actions to meet contextually-determined constraints.

In particular, we address the problem of selecting among a set of alternative reasoning activities (the *control* problem) in service of some object-level problem (the *primary* problem). The control problem exhibits considerable uncertainty since the performance of alternative reasoning methods on the primary problem is highly uncertain. Complex tradeoffs concerning the costs of using alternative methods and directly acting in the world need to be considered . The control problem is essentially an issue of belief management: the presence of significant amounts of uncertainty in a realistic task environment forces a system to constantly face a fundamental choice between using its current information to carry out its primary objectives and making efforts via reasoning activities to improve its state-of-information.[1]

Researchers in artificial intelligence have long been interested in the topic of problem-solving control. Some investigators have focused on developing general architectures with features that support explicit reasoning about control of problem-solving actions [3,4,6,8,11]. The primary emphasis has been on mechanisms and representations by which control knowledge might be used. With few exceptions [12], the knowledge itself has been developed heuristically with little emphasis on developing general principles of control. In this work, we will model the control problem in terms of decision making under uncertainty as formalized in decision theory. As indicated above, control involves resource allocation under uncertainty with complex preferences and the need to reason about the cost and quality of information. Representations and tools from decision theory are a promising path for analysis of these problems from a formal basis [13].

This work has been motivated in large part by a desire to incorporate principled control procedures within autonomous real-time systems. In particular, we are extending Schemer-II, a computational architecture that allows embedding various problem-solving elements in an autonomous system designed for operation in a complex, dynamic, environments [7]. Although Schemer-II's design provides a very robust computational framework for applying appropriately chosen techniques for control reasoning, this architecture does not by itself offer any such techniques for making the appropriate choices. In the latter part of the paper we indicate how the analytic techniques and results derived here are being incorporated into the architecture.

## 2 Decision-Theoretic Control

Our approach to selecting among alternative methods for reasoning about a particular problem is a computational version of an idea proposed by Matheson [16] and more recently Nickerson [17] and Lindley [15]. The outputs of a problem solving method are viewed as *information* in a decision-theoretic sense, that is, the outputs of a model are used to update the probability distribution about an event or potential action.

We are concerned with two classes of action: *primary* actions and *modeling* actions. Primary actions involve the system's interface to the external world, e.g. moving an item, opening a valve, or initiating communication. Modeling actions operate on the system's knowledge to produce new conclusions or recommendations. We use the term modeling to capture the set of actions regarding structuring, solving, and interpreting a model of a domain. Alternative methods may be based on different assumptions and require different amounts of data and time to run. The solutions provides may differ in their quality[14], perhaps expressed in terms of different attributes of a solution [12].

The formulation of the control problem in these terms is illustrated in Figure 1. Modeling actions, $m$, are selected from a space of modeling alternatives $\mathcal{M}$. Sequences of primary actions, $d$, are selected from a space of decision alternatives $\mathcal{D}$. The decisions are represented as square nodes in the figure. The overall utility of the control problem is a function of the primary decision ($d$), some uncertain state of the world ($x$), and the cost of using a particular method ($c$). The cost can be thought of as reflecting the (possibly uncertain) time, data, and processor requirements to use a particular problem solving methodology, $m$.

The output of a model is $s$. It is available at the time the primary decision is made (indicated by the arrow from $s$ to $d$ in Figure 1). The output of a model is uncertain. It also is probabilistically dependent on the state of the world: the information $s$ from using method $m$ provides information about the uncertain state of the world. In this sense, a problem-solving methodology acts as a sensor for some unknown quantity. We express this measure of quality of output as the probability distribution, $\Pr(s|x, m, \xi)$, where $\xi$ is the background state of information (or context) where the distribution applies.

---

[1] Cohen [2] has referred to this tradeoff as balancing internal and external action

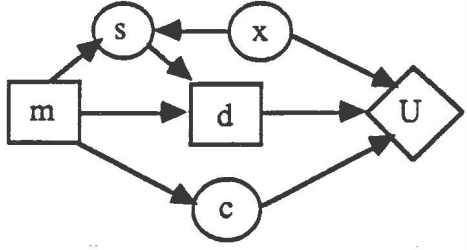

Figure 1: Control-Problem Influence Diagram

The expected utility of the meta-level control problem is

$$\mathrm{E}_{x\,s\,c}(U|d,m,\xi) = \int_c \int_s \int_x U(x,d,c) \Pr(x|s,m,\xi) \Pr(s|m,\xi) \Pr(c|m,\xi)$$

where

$$\Pr(x|s,m,\xi) = \frac{\Pr(s|x,m,\xi)\Pr(x|\xi)}{\int_x \Pr(s|x,m,\xi)\Pr(x|\xi)}$$

by a standard application of Bayes' rule. The optimal primary decision, $d^*(m,s)$ is obtained by solving:

$$\max_{d \in \mathcal{D}} \mathrm{E}_{x\,c}(U|d,m,s,\xi)$$

The distribution over the uncertainty has been updated with the model output. The optimal model $m^*$ is obtained by solving:

$$\max_{m \in \mathcal{M}} \mathrm{E}_{x\,s\,c}(U|d^*(m,s),m,\xi)$$

The above formulation can be extended and operationalized in several ways, as discussed below:

**Resource Usage and Resource Constraints.** In the previous formulation, usage of computational resources is captured in the cost, $c$, of using a particular method. Resources that are limited or that can be expended variably to modify the quality of a computational result can be expressed by conditioning the output, $s$, on the amount of resource available. In a multi-processor system, there is a tradeoff between the amount of time available for a task, and the number of processors assigned to a computation. For example, the quality of an output may depend on time to the next interrupt, $t_i$, and the number of processors, $n$, involving in running method $m$.:

$$\Pr(s|x,m,t_i,n,\xi)$$

In this case $n$ is a control decision while $t_i$ is an uncertain quantity. The form of this distribution can be used to capture behavior of methods whose outputs improve monotonically as additional time or processing power is applied (such as Monte Carlo methods), as opposed to those which require some threshold to provide any useful output.

**Decision Recommendations.** In the previous formulation, the most natural interpretation for the output of a model is that it provides an assessment or diagnosis of some uncertain state of the world. We can modify the formulation for models and methods that provide decision recommendations. As an example, suppose we have an autonomous vehicle that needs to navigate to some objective. The system may embody several alternative means of determining a path to the destination. It could use its logical knowledge to construct a plan while not explicitly considering uncertainty or resource usage. It could develop probability and utility models at various levels of detail, to be solved using exact or approximate methods. It could dispense with a "planning" stage altogether and use local-obstacle-avoidance and reactive-planning methods to attempt to arrive at the destination [9]. Any of these methods will produce output of the same general form: a sequence of actions to be taken, possibly conditional on observations and possibly iterated as a policy (e.g. as in a reactive algorithm).

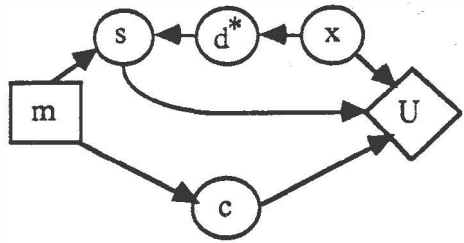

Figure 2: Control Problem Influence Diagram for Decision Recommendations

We illustrate this model with Figure 2. Here we are assuming the existence of some "true" optimal course of action $d^*$ dependent on the state of the world. $d^*$ is the recommendation that would be obtained by maximizing $U_p(x, d)$, the primary decision problem utility function ignoring the costs of reasoning. It is assessed probabilistically, reflecting the system's a priori uncertainty of the optimal action. The output of a method, as previously, provides an estimate of the uncertain optimal decision. The diagram also indicates that the output of the method will be used directly as the primary decision, and the decision model can be solved using Bayes rule and maximization of expected utility in the customary way.

The burden of assessing the probability distribution $Pr(s|d^*, m, \xi)$ can be eased by expressing it in terms of deviations from the optimal value. For example, let $Pr(s = d^*|m, \xi) = p_m$—the probability method $m$ will provide an optimal result is $p_m$. The dispersion about the optimum can be captured in a number of ways depending on the particular method $m$ being considered.[2] A method is *unbiased* for real valued $d^*$ and $s$ when $E(s|d^*, m, \xi) = d^*$.

**Cost of an Error.** A key consideration in the choice among alternative problem solving methods is the extent to which a sub-optimal primary decision will reduce utility in the primary decision problem. For each world state $x$ and alternative $d$ we can calculate $\Delta U = U(x, d^*) - U(x, d)$. These sensitivity measures are indicative of how forgiving a domain is with respect to selection of action. Estimates of this sensitivity can be used to parameterize control strategies across domains and problems, as we would like to perform this type of reasoning without precise specification of $U(x, d, c)$.

We have developed an illustrative example using the control model described above for a specific numerical robot path planning problem [7]. The robot needs to select among a *feasible path method* (**F**), a *basic probabilistic model* (**B**), and a more complex, *information-probabilistic model* (**I**) which considers the possibility of collecting additional information as part of the primary decision problem. Each method is characterized with respect to its probability of providing an optimal solution under uncertainty. The results of the analysis are presented graphically in Figure 3. Optimal regions depend on $t_i$, the time to an interrupt, and , the cost of an error. When $t_i$ and $C_e$ are low then the non-probabilistic modeling method, **F**, is optimal. As these parameters increase, more complete probabilistic reasoning becomes preferred. The type of information summarized in this graph can form the basis for simple control rules, depending on contextual information. Though easy to implement and deliver, they nonetheless are developed based on defensible and clear criteria.

---

[2]Lindley [15] has used assumptions of normality to obtain analytic solutions in a similar problem.

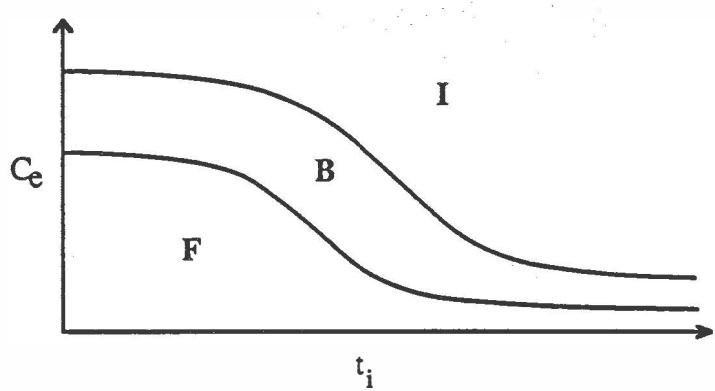

Figure 3: Optimal Reasoning Policies

## 3 The Architecture

Schemer-II is a computational architecture for resource-bounded problem solving.[3] It has been designed to allow for the interleaving of solution-construction, solution-execution, information-gathering, and knowledge-management activities. At a coarse level of description, Schemer-II is an object-based blackboard system. Various problem-solving modules reside in a shared *knowledge space*. The invocation of these modules (or *handlers*) occurs in various ways and is further mediated by the operations of a *top level controller*, which schedules various pending activities for execution, manages communication with the external world, and handles interruption and resumption of ongoing activities.

Schemer-II provides some unique and important features to support flexible, reactive control of problem solving. In particular, the architecture supports a wide variety of techniques for flexible, dynamic scheduling, the ability to employ special-purpose problem-solving modules that can modify the system's control state, and, perhaps most importantly, true pre-emptive control providing the problem-solving system the ability to react promptly and re-focus its attention in response to the occurrence of critical events. However, until recently, both scheduling and pre-emptive control were handled with strictly domain-specific techniques. In this section we use the decision-analytic framework described in Section 2 to analyze: 1) choices amongst alternative problem-solving activities and 2) generation and fielding of interrupt conditions for executing plans.

### 3.1 Scheduling diverse problem-solving elements

The Schemer-II architecture supports encapsulation and interleaved control of multiple, independent problem-solving methods. Schemer-II's handlers, with their object-oriented modularity, meet this requirement by providing a discipline for encapsulating each problem-solving element as a distinct type of object. Each handler can encapsulate a specialized type of problem-solving skill. Handlers provide convenient data structures that support a strong distinction between the information that is strictly local to a problem-solving element and information that is to be shared with other elements.

Handlers in Schemer-II are triggered by changes in data in the system, via communication with external processes, or by direct invocation. In any of these cases, a single triggering event may cause several alternative problem-solving methods to be invoked. Furthermore, at any time multiple tasks may be on the system schedule awaiting execution. The scheduling problem is selection among these alternatives based on computational costs, data requirements, and attributes of the solutions offered by each method.

---

[3]See Fehling [7] for a detailed discussion of the architecture. Successful Schemer applications have been built for a number of real-time, "process management" applications such as diagnosis or control of complex manufacturing processes [3] and automated performance management of advanced avionics systems [10], among others.

In previous implementations the scheduler has used a simple pre-emptive, priority-based scheduling discipline. In this approach the carriers representing potential tasks have an initial fixed priority prescribed by the system developer as a feature of their associated handler. On each cycle of the top level controller the current priority of each task remaining on the schedule is then "aged" (viz., has its priority value modified) in some simple and application-dependent manner.

The disadvantage with this approach is that it is essentially hardwired prior to execution time—there is no general facility to adjust scheduling decisions in response to changes in environmental characteristics. As a supplement to the priority scheme currently in Schemer, we are implementing conflict-detection and -resolution routines for dynamically assigning and updating task priorities. A conflict is detected if several handlers are triggered for execution simultaneously. Once a conflict is detected, the system will look for specialized control knowledge to make a selection or allocation as exemplified by tradeoffs such as in Figure 3. If no specialized knowledge is available or applicable, then a handler which performs decision-theoretic reasoning can examine the conflict, develop a control model such as described in Section 2 and make a recommendation regarding which task should be undertaken.

A potential problem with this approach is entailed by the computational (and other) resource requirements associated with this method of reasoning about control [18,1]. The activities of scheduling are "inner loop" in Schemer-II's overall computational activities. If the computation costs required to explicitly perform a full-blown cost/benefit analysis on each cycle are too high, they will outweigh the value of this control reasoning no matter how formally sound and general it is. Thus, it may be necessary to restrict the real-time estimations performed in scheduling on each cycle in response to limitations such as time deadlines. In extreme cases, it may even be necessary to abandon such a method entirely in favor of the default prioritization scheme.

### 3.2 Critical-event-driven control of reasoning

One of the most important objectives in the evolution of the Schemer-II design has been to fundamentally support problem-solving processes whose control is responsive to critical changes in the problem-solving context. A problem solver dynamically formulating and executing solutions to problems in an uncertain environment must be able to react promptly to the asynchronous occurrence of such critical changes. In response to such changes, the problem-solving system may decide that its current actions are no longer the most preferable ones. In using earlier versions of Schemer in applications that must exhibit reactive, real-time performance, we found the capacity for "interrupt-driven" control of problem-solving to be of paramount importance. In Schemer-II the occurrence of some critical event can initiate a response to immediately interrupt execution of the currently scheduled problem-solving tasks, suspend them gracefully, and commence tasks that are more appropriate in response to the changed information about the problem-solving context. This is readily accomplished by the use of special event-handlers that carry out these actions in response to pre-defined critical events. This aspect of Schemer-II's design is a natural evolution of the mechanisms for "opportunistic control" typical of blackboard systems such as Hearsay-II [5].

The discussion in the previous sections focused entirely on the "planning" phase of a combined control and primary decision problem. A real-time system both plans and executes actions. Suppose the system has solved both the control and primary planning problems, and is now executing the sequence of steps in the primary problem, which may involve a series of compute-intensive low-level tasks. We need to define a set of critical events that would render the current plan inappropriate or inoperable, and signal a need to replan at a higher level.

Critical events are defined with respect to the modeling method used to generate a particular course of action. If the output of model is in the form of a decision recommendation, $d_m^*$, we *annotate* the recommendation with a set of assumptions on which the recommendation was based, as in $(d_m^*, \xi_m)$. The system will continually sense its knowledge base and the environment for conflicts with the set $\xi_m$ and trigger a replanning task when this conflict occurs.

The set $\xi_m$ is a subset of the full set of assumptions (both implicit and explicit) which are embodied in a planning method. There are complex tradeoffs involved in identifying this "critical" subset. Clearly only those assumptions which when violated would cause a change in a recommended action should be included. One class of important assumptions relates to *mutual exclusivity*. If the system detects a condition that is not among a set of enumerated possibilities considered in generating a plan, the plan may be invalid. Other possible classes of critical assumptions relate to the validity of data, probabilities, or defaults used in a model. However, providing sensitivity to critical events with an interrupt structure causes an increase in system overhead and detracts from performance on other tasks. Additional analysis of this tradeoff in the context of real-time reactive planning and execution is needed.

## 4 Conclusion

This paper has described efforts to apply decision analysis to the control of problem-solving within a computational problem-solving architecture. We have addressed control of both assessment and planning methods. The formulation makes it clear that a system's ability to make well-founded decisions about the control of its own problem-solving activities is a problem of information management, and we use concepts based on value of information to perform these allocations.

This research addresses a limitation of much of previous research on problem-solving architectures. Schemer-II is a well tested and highly evolved computational approach to resource-bounded problem solving. The architecture allows encapsulating and interruption of alternative problem-solving methods so that various problem solving techniques can co-exist and be scheduled for execution as needed. One critical limitation of previous research with Schemer was that the methods for coping with uncertainty and for reasoning about control were ad hoc and application-specific. Adoption of decision-theory promises to rectify this shortcoming by providing a set of well-founded and rigorous principles for managing internal resources and other decisions under uncertainty.

Current research efforts involve incorporation of the decision-theoretic control methods described in this paper into the latest implementation of Schemer-II. In future work we will be characterizing various problem solving methods with respect to their quality of information, as well as analyzing time and other resource consumption issues. Additional methods for analyzing interrupt conditions need to be developed, including development of formal justification and generation of interrupts.

## 5 Acknowledgements

We thank Eric Horvitz, Jackie Neider, and Sampath Srinivas for comments on an earlier draft of this paper.